\documentclass[
]{ceurart}

\usepackage{listings}
\begin{document}

\copyrightyear{2025}
\copyrightclause{Copyright for this paper by its authors.
  Use permitted under Creative Commons License Attribution 4.0
  International (CC BY 4.0).}

\conference{CLEF 2025 Working Notes, 9 -- 12 September 2025, Madrid, Spain}


\title{DS@GT at CheckThat! 2025: Ensemble Methods for Detection of Scientific Discourse on Social Media}
\title[mode=sub]{Notebook for the CheckThat! Lab at CLEF 2025}


\author[1]{Ayush Parikh}[
orcid=0009-0002-8364-3011,
email=aparikh49@gatech.edu,
]
\author[1]{Hoang Thanh Thanh Truong}[
orcid=0009-0007-4130-3349,
email=htruong47@gatech.edu
]
\author[1]{Jeanette Schofield}[
orcid=0009-0000-0669-8962,
email=jschofield8@gatech.edu
]
\author[1]{Maximilian Heil}[
orcid=0009-0002-6459-6459,
email=mheil7@gatech.edu
]
\cormark[1]

\address[1]{Georgia Institute of Technology, North Ave NW, Atlanta, GA 30332}
\cortext[1]{Corresponding author.}

\begin{abstract}
    In this paper, we, as the DS@GT team for CLEF 2025 CheckThat! Task 4a Scientific Web Discourse Detection \cite{clef-checkthat:2025:task4}, present the methods we explored for this task. For this multiclass classification task, we determined if a tweet contained a scientific claim, a reference to a scientific study or publication, and/or mentions of scientific entities, such as a university or a scientist. We present 3 modeling approaches for this task: transformer finetuning, few-shot prompting of LLMs, and a combined ensemble model whose design was informed by earlier experiments. Our team placed 7th in the competition, achieving a macro-averaged F1 score of 0.8611, an improvement over the \textit{DeBERTaV3} baseline of 0.8375. Our code is available on Github at \url{https://github.com/dsgt-arc/checkthat-2025-swd/tree/main/subtask-4a}.
\end{abstract}


\maketitle

\section{Introduction}
\hspace{4pt}
Today, a substantial amount of scientific discourse occurs online. Researchers share social media posts announcing their findings, and individuals often contribute to threads discussing the results. Many of these scientific discoveries are discussed on X, a social media platform where posts are referred to as tweets. Building a predictive system that is able to determine the scientific relevance of a tweet can help with fact-checking the tweet for scientific accuracy. Moreover, it can provide researchers with data on mentions (citations) of their work in tweets from others.

CheckThat! 2025 Task 4a \cite{clef-checkthat:2025:task4} asks whether or not tweets contain scientific information. If they do, it then asks whether that information is referring to a scientific claim (Category 1), a reference to a scientific study or publication (Category 2), or a mention of a scientific entity (Category 3).

This paper explores the work the DS@GT team did for Task 4a. We explored transformer fine-tuning and zero-shot, and few-shot prompting using LLMs. We achieved a macro-average F1 score of 0.86, beating the \textit{DeBERTaV3} baseline of 0.84 for the development set. Our submission ranked 6th out of 11 teams based on the macro-average F1 score. Specifically, we placed 7th for Category 1 (Scientific Claims), 5th for Category 2 (Scientific References), and 4th for Category 3 (Scientific Entities). Our implementation is publicly available at \url{https://github.com/dsgt-arc/checkthat-2025-swd/tree/main/subtask-4a}

\section{Related Work}

\subsection{Scientific Discourse Detection}

\hspace{4pt} Prior work in detecting scientific discourse online broadly spans claim detection, entity recognition, citation identification, and scientific fact-checking. 

The task of identifying scientific claims is closely related to stance detection and factuality classification. Earlier research has leveraged large-scale pretrained language models, such as BERT and RoBERTa, fine-tuned on claim verification datasets such as the FEVER \cite{Thorne18Fever} dataset, to identify factual assertions in unstructured text. These approaches often incorporate syntactic features or leverage external knowledge bases.

Entity recognition in social media text has traditionally focused on named entity recognition (NER) using models adapted to noisy and informal language. Scientific entity recognition introduces an additional layer of complexity, as it requires disambiguating institutions, researchers, and domain-specific terminology. Work on domain-specific NER, particularly in biomedical and academic corpora, such as the SciSpacy package \cite{neumann-etal-2019-scispacy}, has informed methods that are increasingly applied to social media platforms. The identification of scientific references, posts that mention or link to scientific studies, has been advanced by research on citation intent classification and altmetrics. This line of work includes approaches to linking social media content to formal publications via DOIs or preprint repositories such as arXiv and bioRxiv.

\subsection{SciTweets}

\hspace{4pt} A key contribution in this space is the SciTweets dataset and annotation framework \cite{SciTweets}. SciTweets offers a structured corpus of annotated tweets, categorized according to their inclusion of scientific claims, references to scientific publications, and mentions of scientific entities. The dataset supports multi-label classification and was created with a detailed annotation schema informed by both academic and public science communication practices. It serves as a valuable benchmark for developing and evaluating models for scientific discourse detection, especially in the context of Subtask 4A for CLEF.

The SciTweets dataset is derived from the TweetsCOV19 dataset \cite{dimitrov:tweetsCOV19}, which is a subset of the TweetsKB dataset \cite{fafalios2018tweetskb}. TweetsKB is a large-scale dataset containing over 1.5 billion English-language tweets collected between 2013 and 2020. TweetsCOV19 filters this collection to focus specifically on COVID-19-related conversations. Building on this valuable resource, SciTweets further refines the data by identifying and annotating tweets that contain scientific claims, reference scientific studies, or mention scientific entities.

\subsection{Evaluation Metrics}

\hspace{4pt} The macro-averaged F1 score is the official metric for CheckThat! Task 4A \cite{clef-checkthat:2025:task4}. The macro-average F1 score is computed as follows:

\begin{equation}
\mathrm{Macro\text{-}F1} = \frac{1}{C} \sum_{i=1}^{C} \mathrm{F1}_i
\label{eq:macro_f1}
\end{equation}

Where C is the total number of categories—in our case, C = 3. The F1 score for each category is shown below:

\begin{equation}
\mathrm{F1} = \frac{2 \cdot \mathrm{Precision} \cdot \mathrm{Recall}}{\mathrm{Precision} + \mathrm{Recall}}
\label{eq:f1_score}
\end{equation}

The macro-averaged F1 score metric is well-suited for multi-label classification tasks with class imbalance, as it assigns equal weight to each category regardless of its frequency. The F1 score has been consistently used across multiple CheckThat! labs over the years, reflecting its importance in capturing both precision and recall in classification tasks.

\subsection{Transformer Fine-Tuning vs. LLM Prompting Approaches}

\hspace{4pt} In this evolving landscape of multi-label classification tasks, two main approaches have emerged: transformer-based fine-tuning and prompting using large language models (LLMs). 

On one hand, fine-tuning methods adapt BERT-based models to domain-specific data. For example, SciTweets employs a SciBERT-based classifier to detect scientific conversation \cite{SciTweets}, demonstrating the effectiveness of tailoring pre-trained models to specialized domains.

On the other hand, prompting-based methods leverage LLMs to perform tasks on natural language instructions. An example of this approach is how \textit{GPT-3} can classify the sentiment of a sentence when provided with a few labeled examples in the prompt \cite{brown2020language}. Without any fine-tuning, the model can perform well on many classification benchmarks, such as sentiment analysis and topic classification. However, despite achieving competitive results, its performance is generally limited compared to transformer models that are fine-tuned on task-specific data.

Recent studies have explored the effectiveness of these approaches for multi-label classification. While prompt-based methods using LLM demonstrated promising results, fine-tuned transformer models(e.g. \textit{RoBERTa} and \textit{DeBERTa}) continue to outperform prompt-based approaches (e.g. \textit{GPT-3.5}, \textit{GPT-4}, and \textit{Claude Opus}) on text classification tasks \cite{bucher2024fine} \cite{bosley2023bert}. 

Nevertheless, LLMs remain widely used due to their generalization capabilities and flexibility in zero-shot and few-shot settings. Their ability to perform a wide range of tasks without task-specific fine-tuning makes them appealing in scenarios where labeled data is scarce and task definitions evolve rapidly, particularly in dynamic settings such as social media discourse on trending scientific topics.

Furthermore, researchers have speculated that LLMs may have a relative advantage in identifying scientific references due to their exposure to citation patterns during pretraining \cite{bucher2024fine}. The study suggests that generative models trained on broad web corpora could be better equipped to recognize citation formats and reference markers. However, this remains a theoretical assumption rather than an empirically validated finding, as the study does not present experimental evidence demonstrating the superior performance of LLM in scientific reference detection tasks.

\subsection{Our Hypothesis}

\hspace{4pt} While prior work suggests that fine-tuned transformer models generally outperform prompt-based LLMs in text classification tasks, it remains unclear how these approaches compare in domain-specific settings. Our research question is whether fine-tuned models may outperform prompt-based LLMs specifically in classifying scientific discourse on social media platforms.
We hypothesize that while fine-tuned models are likely to achieve higher performance on macro-averaged F1 score, prompt-based LLMs may demonstrate superior performance in specific categories, particularly in identifying scientific references (Category 2). This may be due to their broad exposure on web-scale corpora that include citation patterns and academic content. To leverage the strengths of both approaches, we also consider the potential of a hybrid framework that combines fine-tuning and prompting.

\section{Exploratory Data Analysis}

\subsection{Dataset Description}

\hspace{4pt} The training dataset is provided as a TSV file with the following columns:
\begin{enumerate}
    \item index - numerical index for the data samples
    \item text - the text of the tweet
    \item labels - a list of 3 labels, one for each category
\end{enumerate}

There is also a development dataset provided as a TSV file, which also includes ground truth labels. We will report test accuracy on this development set. Lastly, there is an evaluation dataset that only includes the index column and the tweet column with no labels.

The training and development datasets are labeled with binary class labels for 3 separate categories. Category 1 represents whether a tweet contains a scientific claim. Category 2 represents whether a tweet contains a reference to a scientific study or publication. Category 3 represents whether a tweet mentions scientific entities such as a university or a scientist. Note that any further references to Categories 1, 2, and 3 in this paper refer to the classification categories as discussed in this paragraph.

\subsection{Label Distribution}

\begin{figure}[h]
    \centering
    \includegraphics[width=250pt, height= 200pt]{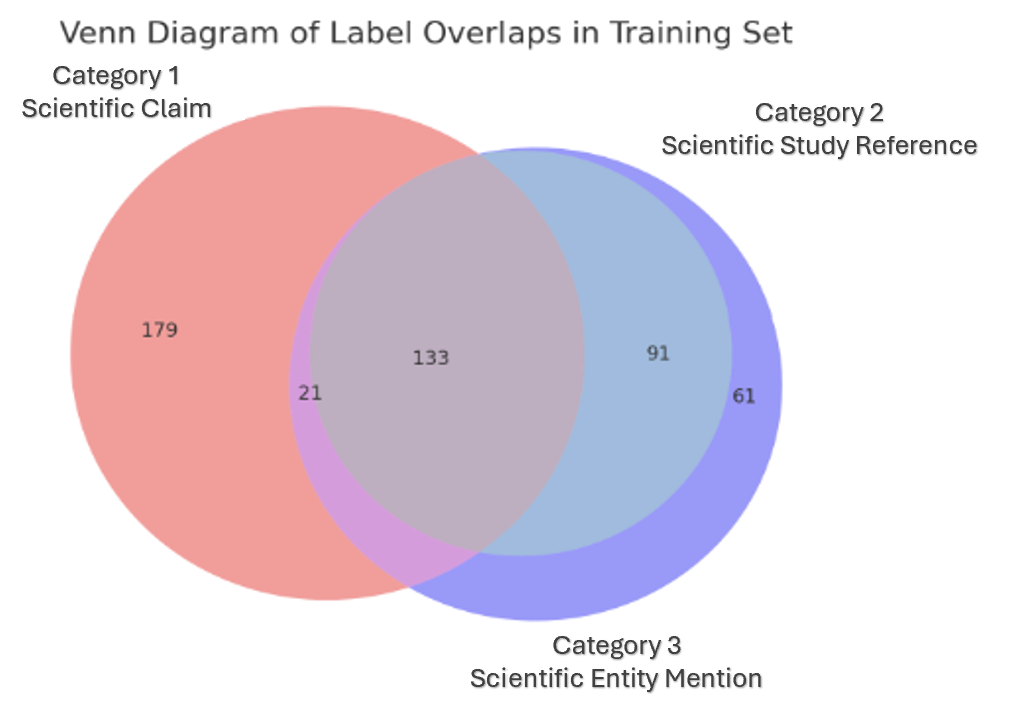}
    \caption{Label overlap in Training Dataset.}
    \label{fig:train_venn}
\end{figure}

\begin{figure}[h]
    \centering
    \includegraphics[width=250pt, height= 200pt]{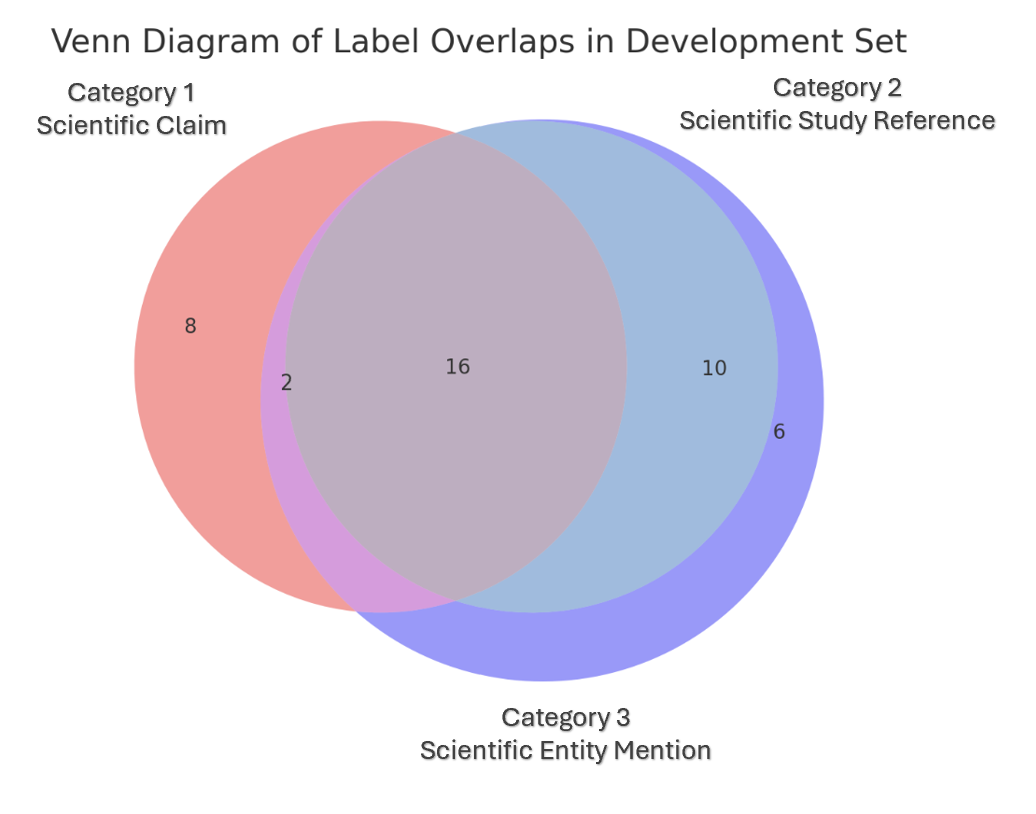}
    \caption{Label overlap in Development Dataset.}
    \label{fig:dev_venn}
\end{figure}

\hspace{4pt} To gain insights into the label distribution and potential class imbalance, we conducted an exploratory data analysis on both the training (ct\_train.tsv) and development (ct\_dev.tsv) datasets. Both datasets consisted of 3 columns: index, text, and labels. Each tweet was annotated with three binary labels representing a multi-label classification task: Category 1 - Scientific Claim, Category 2 - Scientific Study Reference, and Category 3 - Scientific Entity Mention. Each label indicated the presence (1.0) or absence (0.0) of the respective category. 

The training set contains 1,229 tweets, with 333 tweets (27.1\%) labeled as Scientific Claim, 224 (18.2\%) as Scientific Reference, and 306 (24.9\%) as Scientific Entity Mention. The development set contains 137 tweets, with 26 (19.0\%) labeled as Scientific Claim, 26 (19.0\%) as Scientific Reference, and 34 (24.8\%) as Scientific Entity Mention. Furthermore, 736 tweets (59.9\%) in the training set and 81 tweets (59.1\%) in the development set are not labeled with any class, meaning none of the three categories are marked as present. Given that Category 2 has the fewest labeled tweets in both datasets, it may benefit from leveraging the generalization capabilities of large language models (LLMs).

\hspace{4pt} Figures~\ref{fig:train_venn} and~\ref{fig:dev_venn} visualize the overlap between the three classes in the training and development datasets, respectively. In both datasets, we observed a strong overlap between Category 2 - Scientific Study Reference and Category 3 - Scientific Entity Mention. Tweets labeled with Category 2 were also labeled with Category 3. However, the reverse is not always true: There were tweets in Category 3 that do not belong in Category 2. This pattern suggests that Category 2 may represent a more specific subset within the broader scope of Category 3.

\section{Methodology}
\subsection{Overview}

\hspace{4pt} Our approach to the multi-label classification of scientific claims combined the strengths of fine-tuned transformer models and LLMs.

\hspace{4pt} We experimented with various transformer encoders using a \textit{DeBERTa}-based model as the baseline, as it was provided by the organizers as the official baseline for the competition \cite{clef-checkthat:2025:task4}. The model achieving the highest macro-averaged F1 score on the development set was selected for fine-tuning. Using the training dataset, the model was trained on labeled tweet data to predict three independent binary categories. We decided to experiment with transformers as they have been shown in the past to perform well on this task, as indicated by our baseline \cite{clef-checkthat:2025:task4}. We leveraged the scikit-learn and PyTorch packages to help finetune and test transformer models \cite{scikit-learn} \cite{paszke2019pytorchimperativestylehighperformance}.

\hspace{4pt} On the other hand, we explored prompting LLMs using both zero-shot and few-shot strategies to classify tweets based on natural language descriptions of the categories. LLMs have been shown to generalize well to a wide variety of tasks, including fact-checking \cite{vykopal2024generativelargelanguagemodels}. They are also trained on practically the entire Internet, making them powerful tools that generalize well to other tasks \cite{karpathy202deep}. As a result, we decided to experiment with LLMs for this task. In particular, we experimented with \textit{GPT-4o-mini} and \textit{GPT-4o} using both zero-shot and few-shot prompting strategies. In the end, we adopted the few-shot prompting strategy with semantically retrieved examples using \textit{GPT-4o} in our final pipeline, as it achieved the highest macro-average F1 score on the development set. See the Results section for more details. We leveraged LangChain and scikit-learn to in order to collect and analyze outputs from the LLMs \cite{scikit-learn} \cite{langchain2022}.

\begin{figure}
    \centering
    \includegraphics[width=1.0\linewidth]{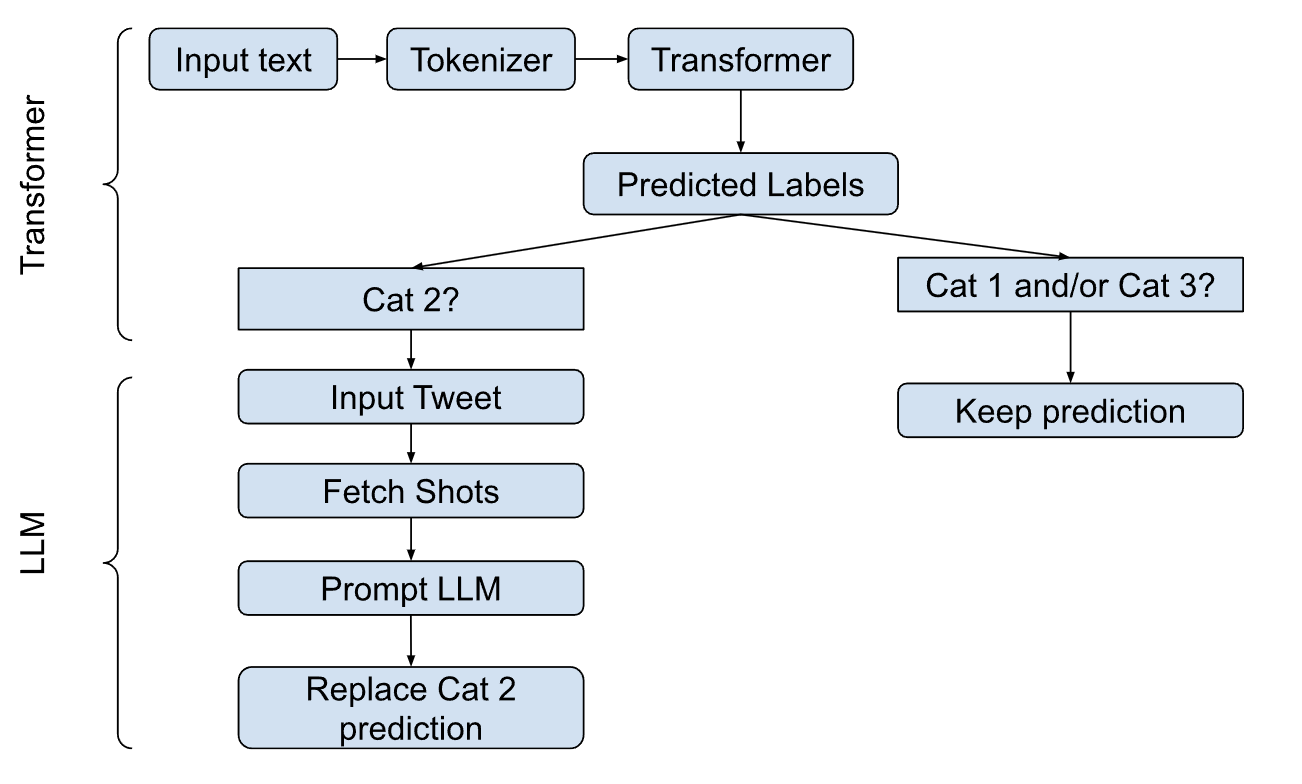}
    \caption{Classification pipeline for our finalized ensemble model.}
    \label{fig:classification-pipeline}
\end{figure}

\hspace{4pt} Observing that the fine-tuned transformer performed better on Categories 1 and 3, while LLMs yielded better results at Category 2, we developed a combined ensemble model. As illustrated in Figure~\ref{fig:classification-pipeline}, our final classification pipeline integrates both fine-tuned transformer and LLM predictions in an ensemble framework. The transformer model handles predictions for Category 1 (Scientific Claim) and Category 3 (Scientific Entity Mention), while the LLM is responsible for Category 2 (Scientific Study Reference) using few-shot prompting with semantically retrieved examples.

\subsection{Transformer Finetuning}
\hspace{4pt} We explored different transformer models and selected one with the highest macro-average F1 score on the development set. We fine-tuned the pre-trained transformer encoder \textit{microsoft/deberta-v3-base} on the training dataset. The data was split into a 90/10 ratio for training and validation, respectively. Each tweet was tokenized using the same tokenizer as the base model to ensure consistency with the underlying architecture. The tokenized inputs were then passed through the transformer, and the resulting embeddings were fed into a feedforward neural network to predict the three categories. 

\hspace{4pt} The model produced raw logits, one for each category, which were then passed through a sigmoid activation to generate the probabilities of three categories. If the sigmoid probability exceeded a threshold of 0.5, the tweet was classified as belonging to that label. The model was learned by comparing its predictions to the correct labels using Binary Cross-Entropy with Logits Loss (BCEWithLogitsLoss), which is well-suited for the nature of multi-label classification tasks. 

\hspace{4pt} We trained the model for up to 20 epochs, applying early stopping if validation performance did not improve for three consecutive epochs. The best-performing model, based on the macro F1 score on the validation set, was saved.

\subsection{LLM Approaches}
\label{sec:llm_approaches}
\hspace{4pt} Another approach we took for this multiclass classification task is to leverage the generalizability of LLMs. LLMs have gained popularity in recent years. As these models have grown in size, they have also gotten better at several tasks, including question-answering. As a result, we decided to experiment with LLMs for this multiclass classification task by constructing a custom prompt asking the LLM to identify which of the 3 classes is applicable. In particular, we experimented with both zero-shot and few-shot prompting.

The zero-shot prompt we used is as follows:
\begin{quote}
\ttfamily
You are a helpful assistant who classifies tweets into 0 or more categories.
The categories are: 1. Contains a scientific claim. 2. Refers to a scientific study/publication. 3. Mentions scientific entities (e.g., a university or scientist). 4. None of the above. You need to provide all applicable category numbers as a one-hot encoded list of size 3 (e.g., [1.0, 1.0, 0], [1.0, 0, 0], [0.0, 0.0, 1.0]). If the tweet does not fit into any category, return [0.0, 0.0, 0.0]. You must always return a list of 3 elements as such. Classify the following tweet into one or more of the following categories: 1. Contains a scientific claim. 2. Refers to a scientific study/publication. 3. Mentions scientific entities (e.g., a university or scientist). 4. None of the above. Provide all applicable category numbers as a one-hot encoded list of size 3 (e.g., [1.0, 1.0, 0], [1.0, 0, 0], [0.0, 0.0, 1.0]). If the tweet does not fit into any category, return [0.0, 0.0, 0.0]. You must always return a list of 3 elements as such. Tweet: {tweet}.
\end{quote}
We also experimented with few-shot prompting. In particular, we used the SemanticSimilarityExampleSelector with a FAISS vector store as provided by LangChain in order to select training set examples to use as shots. Given a new tweet to produce a set of multiclass labels for, the example selector would look up 5 similar examples from its database of tweets from the training set to use as shots within the LLM prompt. In this way, the prompt would read similarly to the zero-shot prompt above, with the following additional text:
\begin{quote}
\ttfamily
Here are some example tweets along with their classifications:
\end{quote}
Each example tweet would be appended at the end of this prompt.

\subsection{Combined Model}
\hspace{4pt} After recognizing that the LLM approach was better at Category 2 predictions, whereas the finetuned BERT model was better at Categories 1 and 3, we converged on a combined ensemble model. Figure \ref{fig:classification-pipeline} depicts the details of our final classification pipeline. The system takes a tweet as input. It tokenizes the input text and passes these tokens to our finetuned transformer model. The transformer outputs predictions for the 3 categories. In the case of categories 1 and 3, we retain the transformer's predicted labels. In the case of category 2, we throw out the prediction and instead, take the original input tweet text to look up 5 shots (example tweets) with the closest semantic similarity to the input tweet. Then, we use the few-shot prompt discussed in Section 3.4 to prompt the LLM. Finally, we take the category 2 prediction provided by the LLM to produce a final set of 3 predictions.

\section{Results}

\begin{table}[h!]
\centering
\caption{Performance comparison of various models across macro-averaged and category-wise F1 scores}
\begin{tabular}{@{}lcccc@{}}
\toprule
\textbf{Model} & \textbf{Macro-avg F1} & \textbf{Cat1 F1} & \textbf{Cat2 F1} & \textbf{Cat3 F1} \\
\midrule
DeBERTaV3-Baseline & 0.84 & 0.82 & 0.79 & 0.90 \\
Fine-tuned microsoft-deberta-v3-base & 0.85 & 0.86 & 0.82 & 0.87 \\
microsoft-deberta-v3-large & 0.85 & 0.83 & 0.86 & 0.87 \\
Fine-tuned microsoft-deberta-v3-large & 0.80 & 0.85 & 0.74 & 0.83 \\
MarieAngeA13-Sentiment-Analysis-BERT & 0.77 & 0.75 & 0.74 & 0.81 \\
GPT-4o-mini (Zero Shot) & 0.75 & 0.78 & 0.86 & 0.62 \\
GPT-4o-mini (5-shot, semantic sim) & 0.77 & 0.77 & 0.90 & 0.65 \\
GPT-4o (5-shot, semantic sim) & 0.81 & 0.77 & 0.89 & 0.77 \\
\textbf{Combined DeBERTA + LLM (Cat2)} & 0.86 & 0.86 & 0.85 & 0.87 \\
\bottomrule
\end{tabular}
\label{tab:model-comparison}
\end{table}

\subsection{Baseline model}
\hspace{4pt} Table \ref{tab:model-comparison} summarizes our results on the development set. The baseline model demonstrated strong performance across all three categories with a macro-average F1 score of 0.84. Category 3 performed the best overall, with an F1 score of 0.90, followed by Category 1 with 0.82, and Category 2 with 0.79. These results indicate that while the baseline model performs robustly across all three categories, classification for Category 2 shows room for improvement, potentially due to its lower representation in the dataset. We could leverage LLMs's capacity in generalization to provide additional support in this category.

\subsection{DeBERTa + LLM Ensemble}
\hspace{4pt} Among all models, the best-performing approach is the combined ensemble of the \textit{DeBERTa} and LLM models, which achieved a macro-averaged F1 score of 0.86. This ensemble outperformed the baseline on Categories 1 and 2, with F1 scores of 0.86 and 0.85, respectively. However, it performed slightly worse on Category 3, achieving an F1 score of 0.87 compared to the baseline's 0.90. This result suggests that leveraging an LLM for Category 2, where data representation was lower, helped improve overall performance as reflected in the F1 score.

\subsection{Transformer models}

\hspace{4pt} Among the pretrained transformer models, \textit{microsoft-deberta-v3-large} achieved the highest macro-averaged F1 score (0.85), followed by \textit{microsoft-deberta-v3-base} (0.84) and \textit{MarieAngeA13} (0.77). 

Due to the promising results of both \textit{DeBERTa} models on the development set, we proceeded to fine-tune them. Interestingly, the fine-tuned \textit{DeBERTa-base} model improved slightly, increasing its macro-averaged F1 score from 0.84 to 0.85. In contrast, the fine-tuned \textit{DeBERTa-large} model saw a decrease in performance, with its F1 score dropping from 0.85 to 0.80. This result may suggest that the large model overfitted to the training data, which limited its ability to generalize effectively to the development set. Because the fine-tuned \textit{DeBERTa} base model achieved the highest F1-score in Category 1 and Category 3, we decided to use this model in our ensemble model.

\subsection{LLM models}

\hspace{4pt} The \textit{GPT-4o-mini} (zero-shot), \textit{GPT-4o-mini} (5-shot), and \textit{GPT-4o} (5-shot) models all performed worse than the baseline, with macro-averaged F1 scores of 0.75, 0.77, and 0.81, respectively. However, all \textit{GPT-4o} models outperformed the baseline in Category 2, with GPT-4o-mini (few-shot) achieving the highest F1 score of 0.90, the best among all models for this category. The fact that LLM models perform best in Category 2—despite it having the smallest amount of data—suggests that LLMs are particularly effective at identifying scientific references, likely due to their exposure to web-based content and citation patterns. You might be wondering why the amount of data would even affect LLM performance. The argument there is that since we are employing few-shot prompting, having a larger sample set of data to use as shots provides more variety for the LLM to utilize. This increases the likelihood that similar tweets exist in the corpus. Thus, high performance in Category 2 despite having fewer examples is an interesting result that demonstrates the LLM's inherent deep understanding of scientific references and their structure across the web.

\section{Discussion}

\hspace{4pt} Here, we discuss our results and their implications. In particular, we closely examine why transformer finetuning performed better on Categories 1 and 3 while the LLM fared better on Category 2.

\subsection{Finetuning Performs Best for Categories 1 and 3}
\begin{figure}[h]
    \centering
    \includegraphics[width=0.9\linewidth]{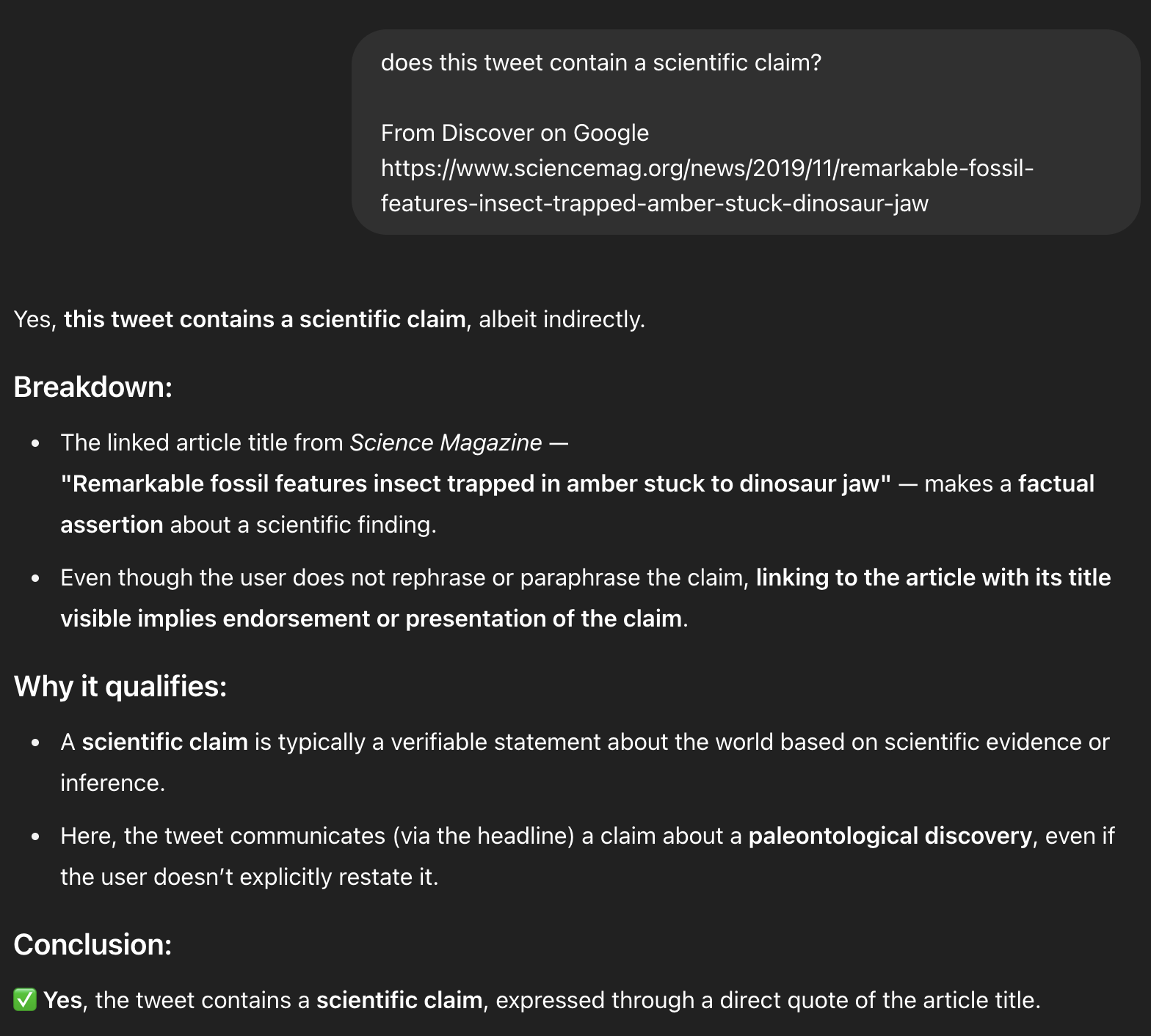}
    \caption{GPT-4o gets confused for classifying some tweets into Category 3.}
    \label{fig:Cat3-Confusion}
\end{figure}

\hspace{4pt} Finetuning transformer models performs best for category 1, determining if a tweet contains a scientific claim, as well as for category 3, determining if a tweet mentions scientific entities such as a university or scientist. The LLM is susceptible to confusion when trying to make determinations for these two categories. As shown in Figure \ref{fig:Cat3-Confusion}, the LLM gets confused about whether the tweet contains a scientific claim. Please note that here, we use a subset of the entire prompt detailed in \ref{sec:llm_approaches} to specifically understand potential sources of LLM confusion for category 2. The finetuned transformer correctly identifies that this tweet does NOT contain a scientific claim, but the LLM thinks it indirectly does because it references an article whose headline mentions a discovery. The LLM is overthinking this. It has seen too much content for its good, leading it to go down the wrong path. Note that LLMs are inherently non-deterministic and may yield different results for the same prompt, but through this example, we can see one example of how confusion may lead to incorrect classification.


\subsection{LLM Performs Well for Category 2}
\begin{figure}[h]
    \centering
    \includegraphics[width=0.9\linewidth]{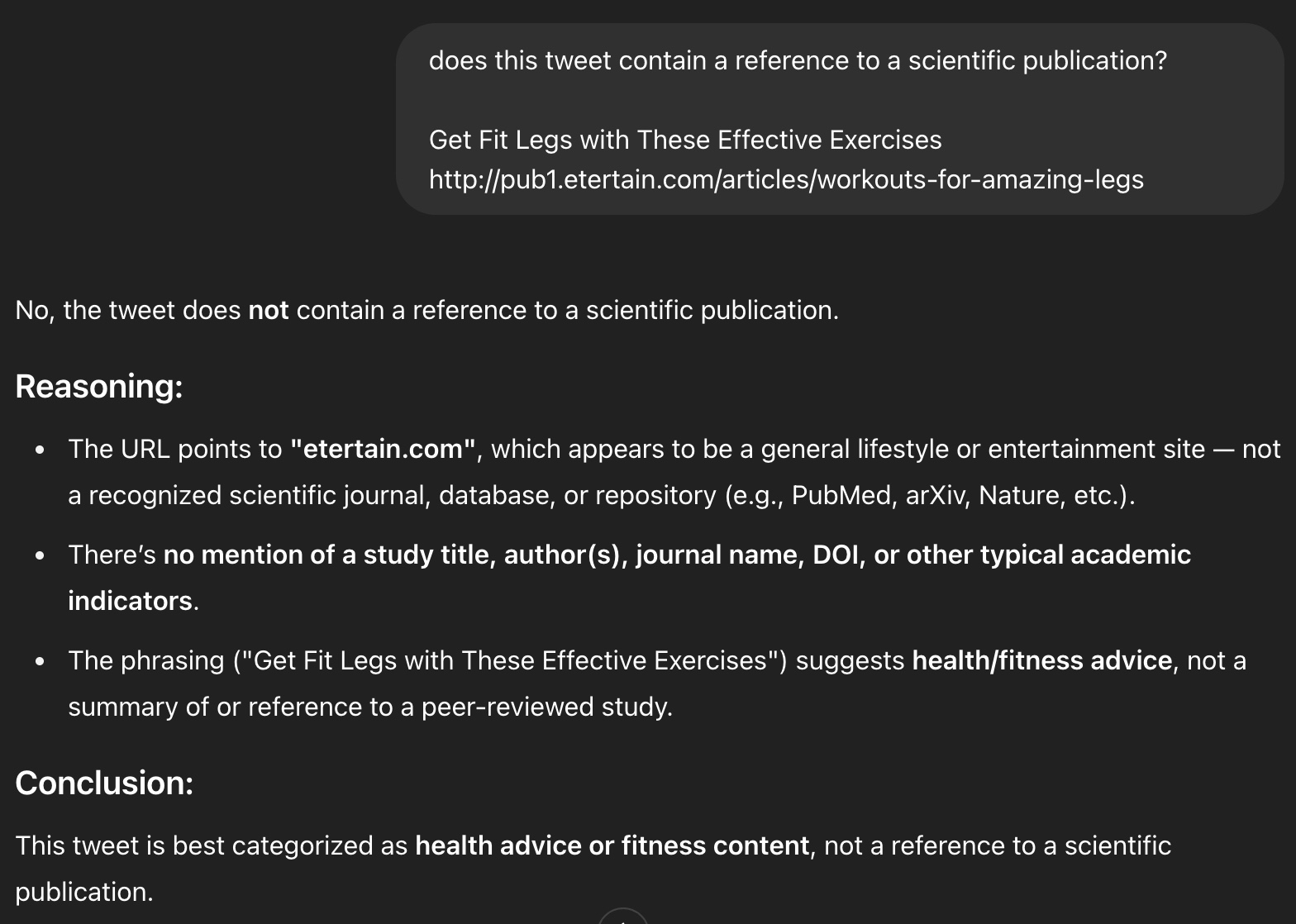}
    \caption{GPT 4o performs well when classifying tweets into Category 2.}
    \label{fig:Cat2-Success}
\end{figure}

\hspace{4pt} LLMs perform extremely well on category 2, namely determining if a tweet contains a reference to a scientific study or publication. One likely reason for this performance improvement is the simple fact that LLMs are trained on the entirety of the internet and thus have an inherent ability to understand links well. As Andrej Karpathy discusses in \cite{karpathy202deep}, the Internet is filled with links, and LLMs are often absorbing information about how these links are structured. Even with zero shots, \textit{GPT-4o-mini} performs well on Category 2 with an F1 score of 0.86, far better than its performance on other categories. This suggests the LLM may have some intrinsic knowledge about how scientific links look compared to others. This is further confirmed when we took a closer look at examples that \textit{GPT-4o} predicted correctly for Category 2 but were predicted incorrectly by the finetuned transformer model. As shown by Figure \ref{fig:Cat2-Success}, the LLM can break down why a link is not referencing a scientific publication in great detail. The LLM understands what typical scientific links look like. While Categories 1 and 3 require some additional factual knowledge and verification, the LLM may simply be able to game Category 2 performance based on the billions of links in its training data. 


\section{Future Work}
\hspace{4pt} There are several ideas we could explore further in an attempt to improve model performance. We will first discuss possible improvements for model fine-tuning, followed by potential enhancements to the LLM approach.

\subsection{Model Finetuning Future Directions}
\hspace{4pt} To improve the performance of transformer models, we would take into account the dependencies of Categories 2 and 3 in our pipeline. As mentioned in our exploratory data analysis, tweets in Category 2 were always labeled with Category 3, suggesting a strong directional correlation between the two. However, our current approach, treating each category as an independent binary classification task, fails to capture this dependency. To address this issue, we will implement a new pipeline that explicitly models the dependency between categories. This strategy enables the model to better exploit underlying data patterns, potentially improving its classification performance.

\subsection{LLM Approach Future Directions}
\hspace{4pt} There are several additional avenues we could explore to try to improve the performance of our LLM approach. First and foremost, we could further experiment with different LLMs. In particular, we could try the latest LLama models, namely LLama 4 Scout and Maverick, to see how they fare against the OpenAI models we tried. We could also try the latest \textit{Claude Sonnet} model for comparison. Beyond this, we could experiment with tool use by allowing models to search the web. As shown in \cite{langchain_tool_calling}, LangChain, the package we used to prompt \textit{GPT-4o}, does not support tool calls out of the box. We would need additional code and setup to allow the system to browse the way. By leveraging tool calling, we could allow our system to look up whether a university exists or whether a person mentioned in the tweet is an actual scientist. We could even further validate whether a study mentioned in the tweet is real and covers a scientific topic.

\hspace{4pt} Beyond enabling tool calling, we could also experiment with different prompting strategies. One approach might be to be more explicit in the prompt. The LLM approach fared relatively poorly in Category 3. This is likely because we were not explicit enough in the prompt about what exactly falls under consideration for a scientific entity \textit{besides} a university or a scientist. Another approach might be to split up prompts for each category and combine the 3 results at the end. In this way, the LLM would be solely focused on one category at a time and would treat each separately, potentially avoiding confusion.



\section{Conclusions}

\hspace{4pt} In this paper, we discuss 3 approaches for Subtask 4a for CLEF 2025 related to scientific web discourse detection. In particular, we discussed model finetuning, LLM approaches, and a combined model that took the best of both worlds. Our best model beats the baseline \textit{DeBERTaV3} model on macro-average F1 score with a score of 0.86 versus 0.84 for the baseline on the set. We also beat the baseline on categories 1 and 2 as well with F1 scores of 0.86 and 0.85, respectively. 

\section*{Acknowledgements}
\hspace{4pt} Thank you to the DS@GT CLEF team for their support. Special thanks to Anthony Miyaguchi and Murilo Gustineli for their support and for leading the DS@GT CLEF research group. Thank you to Partnership for an Advanced Computing Environment (PACE) \cite{PACE} at the Georgia Institute of Technology, Atlanta, Georgia, USA, for allowing us to use their resources to perform this research.

\section*{Declaration on Generative AI}
\hspace{4pt} During the preparation of this work, the authors used \textit{GPT-4o} in order to \textbf{draft content}. In particular, the authors leveraged AI assistance to help draft content for the Related Work section. The authors also used \textit{GPT-4o} for \textbf{citation management} to help correctly structure citations in BibTeX format. After using these tools, the authors reviewed and edited the content as needed and took full responsibility for the publication’s content.

\bibliography{main}

\end{document}